\documentclass{article}
\usepackage{iclr2027_conference,times}

\usepackage{amsmath,amssymb}

\usepackage[T1]{fontenc}
\usepackage{hyperref}
\usepackage{booktabs}
\usepackage{multirow}
\usepackage{microtype}
\usepackage{xcolor}
\usepackage{colortbl}
\usepackage{graphicx}
\usepackage{float}
\usepackage{wrapfig}
\usepackage{tikz}
\usepackage[capitalize,noabbrev,nameinlink]{cleveref}
\hypersetup{
  hidelinks,
  pdftitle={GeoScaffold: Learning Compact Geometric Latents via Reconstruction for Efficient Vision-Language Navigation},
  pdfauthor={Yixuan Jiang, Wentong Li, An Liu, Zihao Xin, Fulin Tang, Cong Leng, Yang Gao, Jian Cheng}
}

\makeatletter
\def\paragraph{\@startsection{paragraph}{4}{\z@}{0.7ex plus 0.2ex minus .2ex}{-1em}{\normalsize\bf}}
\makeatother

\newcommand{\R}{\mathbb{R}}
\DeclareMathOperator{\softmax}{softmax}

\newcommand{\geotok}[1]{\langle\mathrm{geo}_{#1}\rangle}
\newcommand{\numq}{K}
\newcommand{\numtok}{N}
\newcommand{\numcode}{C}
\newcommand{\vqenc}{\mathcal{E}}
\newcommand{\vqdec}{\mathcal{D}}
\newcommand{\codebook}{\mathbf{B}}
\newcommand{\lact}{\mathcal{L}_{\mathrm{policy}}}
\newcommand{\ltok}{\mathcal{L}_{\mathrm{tok}}}
\newcommand{\lrec}{\mathcal{L}_{\mathrm{rec}}}
\newcommand{\dmodel}{d}
\newcommand{\depthmap}{D}

\title{GeoScaffold: Learning Compact Geometric\\ Latents via Reconstruction for Efficient\\ Vision-Language Navigation}

\author{Yixuan Jiang$^{1,2}$, Wentong Li$^{3}$, An Liu$^{2}$, Zihao Xin$^{3}$, Fulin Tang$^{2,4}$\thanks{Corresponding author: \texttt{tangfulin2015@ia.ac.cn}.}\hphantom{$^{*}$}, Cong Leng$^{4}$, \\
\textbf{Yang Gao$^{1}$, Jian Cheng$^{2}$} \\
$^{1}$Nanjing University, Nanjing, China \\
$^{2}$Institute of Automation, Chinese Academy of Sciences, Beijing, China \\
$^{3}$Nanjing University of Aeronautics and Astronautics, Nanjing, China \\
$^{4}$MAICRO, Nanjing, China
}

\begin{document}

\maketitle

\begin{abstract}
Recent vision-and-language navigation (VLN) systems increasingly adopt streaming Video-LLM policies that map egocentric RGB observations and instructions directly to low-level actions. Yet these policies inherit weak 3D geometric priors from 2D pretraining. Existing geometry-aware extensions charge a persistent inference-time price: depth sensors, 3D encoders, or per-step perception tool calls. We propose GeoScaffold, a geometric supervision framework that pays this price once, at training time, by internalizing geometry into the policy itself. It first learns a compact depth tokenizer on depth maps from the training trajectories and freezes it. It then fine-tunes the policy with a handful of learnable geometry query tokens, training their hidden states to reconstruct navigation-critical geometry such as depth, connectivity, and traversability. This supervision turns the query states into compact geometric latents for action decoding, and through the shared weights also internalizes geometry into the backbone's own representations. Like a scaffold, the tokenizer, target generators, and reconstruction heads are discarded after training, leaving the backbone and action interface unchanged. Extensive experiments show that GeoScaffold consistently outperforms leading vision-only navigators on continuous VLN benchmarks, offering a practical paradigm for lightweight edge deployment of spatially aware embodied navigation models.

\end{abstract}

\section{Introduction}
\label{sec:intro}
Vision-and-language navigation (VLN) requires an embodied agent to follow natural-language instructions while moving through a previously unseen environment \citep{anderson2018vision,ku2020room,krantz2020beyond}. Recent progress has increasingly moved VLN toward an end-to-end vision-language-action (VLA) paradigm, in which a Video-LLM policy maps egocentric RGB observations and the instruction to navigation actions step by step \citep{zhang2024navid,cheng2024navila,wei2025streamvln,zhang2025navfom,xin2026decovln}. However, the backbones underlying these policies are pretrained almost entirely on internet-scale 2D data, which makes them strong at semantic recognition yet weak at metric and topological reasoning \citep{chen2024spatialvlm,fu2024blink,yang2025thinking}. Navigation demands exactly these abilities. Beyond recognizing objects and matching them to linguistic landmarks, an agent must reason about free space, relative scale, doorway structure, and route connectivity. These policies therefore struggle with depth-sensitive decisions, such as judging whether a landmark is nearby, whether a passage is traversable, or whether to continue exploring rather than stop prematurely.
\begin{figure}[t]
    \centering
    \includegraphics[page=1,width=\textwidth,trim=562 648 544 637,clip]{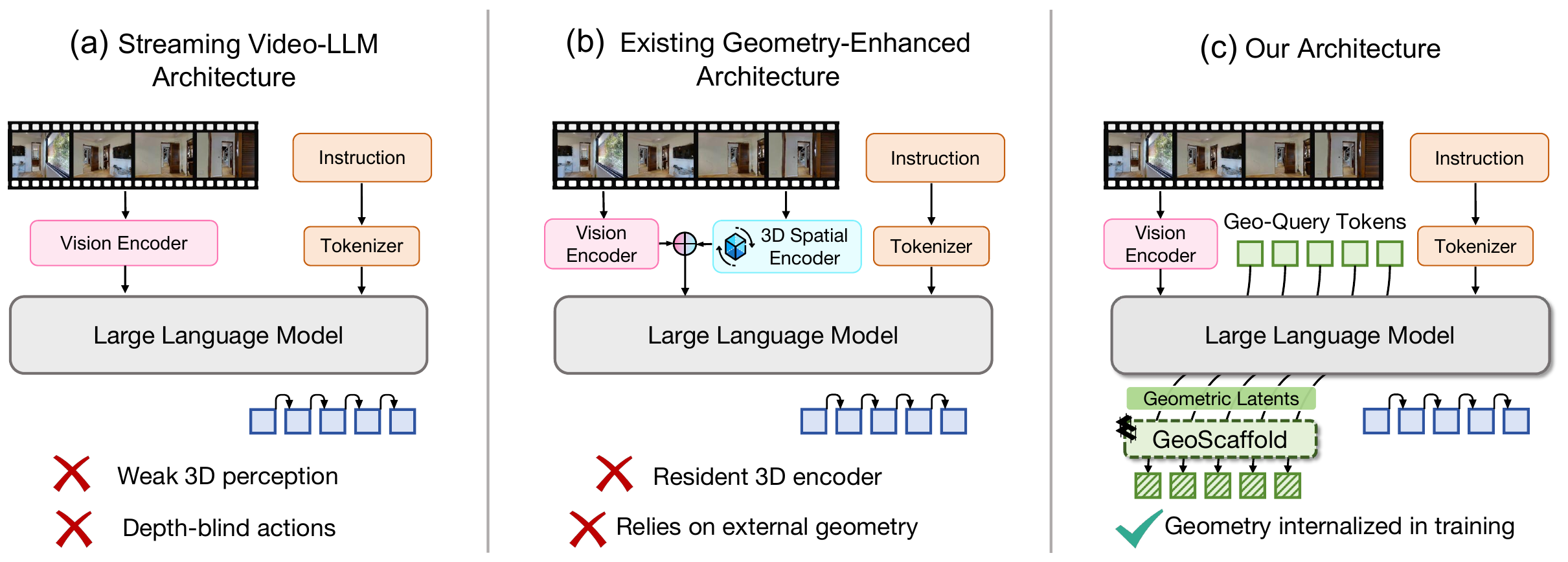}
    \vspace{-14pt}
    \caption{Comparison of end-to-end VLN paradigms. (a)~Streaming Video-LLMs act directly from RGB, but lack 3D perception. (b)~Geometry-enhanced policies add a spatial encoder, but it stays resident at inference and geometry is never internalized. (c)~GeoScaffold internalizes geometry via training-only query-token supervision, leaving the deployed RGB-and-text action path unchanged.}
    \label{fig:second-page-overview}
\end{figure}

Existing methods compensate for this weakness with additional machinery at inference time. Geometry-enhanced methods give the policy explicit spatial support, running a frozen 3D encoder alongside the backbone \citep{zeng2025janusvln,fan2026prospect,liu2026dygeovln} or lifting RGB-D observations into geometry-aware BEV maps \citep{yang2026gavln}. As shown in \cref{fig:second-page-overview}(b), these components remain active at deployment, with depth sensors adding hardware requirements and 3D encoders increasing GPU memory usage and inference latency. Other methods restructure the system, adopting dual-system designs in which a slow reasoning module guides a fast controller \citep{wei2025dualvln,internrobotics2025internvla,xue2026omninav} or zero-shot pipelines of LLM planners, perception, and path-planning components \citep{zhou2024navgpt,long2024instructnav,chen2025aoplanner}. Such designs require coordination among multiple modules, complicating system integration and deployment.

By contrast, humans act with rich spatial awareness from visual experience alone, without explicit depth sensing at every step. For navigation, such spatial awareness should help the agent judge distances, identify traversable regions, and recognize passages that connect spaces. We therefore ask whether geometric supervision can equip an end-to-end VLA navigator with these abilities during training, while requiring only RGB observations and language instructions at deployment.

To this end, we propose \textbf{GeoScaffold}, a training-only geometric enhancement framework for streaming VLN. GeoScaffold supervises the policy with three geometric targets. Depth captures metric scale, while connectivity and traversability mark route-relevant passages and free space, respectively. The supervision enters through a small set of learnable geometry query tokens \citep{carion2020detr,li2023blip2,mu2023gist,lai2024lisa}, inserted before each action block, one dedicated group per target. Each group learns to reconstruct its own target from its hidden states. Training proceeds in two stages. Stage I learns a compact depth tokenizer offline. Stage II then fine-tunes the policy with the navigation objective unchanged plus the auxiliary reconstruction losses. This pressure compresses scene geometry into the query hidden states, which act as compact geometric latents for action decoding. After training, all auxiliary modules are discarded. The deployed policy runs on RGB and text alone, with only the learned geometry tokens remaining in the prompt.

Our contributions are threefold:
\begingroup
\setlength{\topsep}{3pt}
\setlength{\itemsep}{2pt}
\setlength{\parsep}{0pt}
\begin{itemize}
    \item We introduce \textbf{GeoScaffold}, a lightweight, removable geometric enhancement framework for efficient end-to-end streaming VLN that requires no depth sensors, 3D encoders, or reconstruction modules at inference time and deploys on edge hardware.
    \item We design a query-token reconstruction objective whose dedicated depth, connectivity, and traversability groups shape their hidden states into compact geometric latents.
    \item We present an empirical study of training-only geometric supervision for VLN, showing consistent success-rate gains across different backbones and in real-world deployment, with only the learned geometry tokens retained at inference.
\end{itemize}
\endgroup

\section{Related Work}
\label{sec:related}

\subsection{VLA Model for Navigation}
\label{sec:rw_vln}

Recent VLN research has increasingly reformulated navigation as an end-to-end vision-language-action problem. Navigators of this kind fine-tune a multimodal language model to map visual observations and the instruction to navigation decisions \citep{zhang2024navid,cheng2024navila,zheng2024navillm,zhang2025uninavid,zhang2025navfom}, inheriting rich visual priors and the commonsense knowledge of the backbone. To run over long trajectories, other work turned the interaction into a streaming multi-turn dialogue whose cached context and compact memory keep the cost of each step bounded \citep{wei2025streamvln,xin2026decovln}. Building on these policies, dual-system designs add a slow reasoning module above a fast controller \citep{wei2025dualvln,internrobotics2025internvla,xue2026omninav}. However, their backbones are pretrained almost exclusively on 2D web images and videos, so the resulting policies inherit little sense of metric scale or 3D structure, precisely the abilities that navigation requires.

\subsection{Geometry-Enhanced VLN}
\label{sec:rw_geometry}

A broad line of VLN research addresses this spatial weakness by injecting depth or 3D structure into the agent. These methods make geometry an input to the policy at deployment time, whether through RGB-D or stereo observations \citep{liu2025navid4d,wang2026stereonav}, 3D-language feature fields and tokens \citep{wang2025g3dlf,wang2025dynam3d}, map-like memories \citep{wang2023gridmm,zhang2025mapnav,zhang2026spatialnav,yang2026gavln}, frozen 3D foundation models \citep{zeng2025janusvln,fan2026prospect,liu2026dygeovln} such as VGGT \citep{wang2025vggt}, or agentic perception tools that query occupancy, LiDAR, or depth on demand \citep{xin2026agentvln,yue2026spatialvln}. These designs improve spatial understanding, but the sensors, encoders, maps, or tool calls that supply geometry must stay active at deployment and add latency, memory, or hardware to the robot. GeoScaffold instead uses geometry only as a training-time supervision signal, so the policy learns geometric representations from RGB alone and carries them into deployment as compact latents, with no geometry module on the inference path.

\subsection{Auxiliary Perception Supervision and Query Tokens}
\label{sec:rw_query}

Auxiliary perception supervision enriches visual representations through prediction and reconstruction tasks. Early navigation methods explore depth prediction \citep{mirowski2017learning} and self-supervised reasoning \citep{zhu2020auxrn}. Recent approaches extend this idea to multimodal LLMs and VLAs: ROSS \citep{wang2025ross}, 3DRS \citep{huang2025threedrs}, and Spatial Forcing \citep{li2025spatialforcing} use visual reconstruction or 3D feature alignment, while DreamVLA \citep{zhang2025dreamvla} predicts future world knowledge for manipulation. Query-based aggregation and special-token readouts, as in DETR \citep{carion2020detr}, BLIP-2 \citep{li2023blip2}, and LISA \citep{lai2024lisa}, provide compact interfaces for visual prediction. Perception Tokens \citep{bigverdi2025perception} and Perceptio \citep{li2026perceptio} also explore autoregressive perception-token generation. GeoScaffold uses dedicated query states to reconstruct current-frame depth, connectivity, and traversability during training, learning compact geometric latents for RGB-only streaming navigation.

\section{Method}
\label{sec:method}

\subsection{Preliminaries}
\label{sec:prelim}

We study vision-and-language navigation in continuous environments (VLN-CE) \citep{krantz2020beyond}, in which an agent follows a natural-language instruction using egocentric RGB observations and predicts actions from the action space $\mathcal{A}=\{\textsc{Forward}~25\,\mathrm{cm},\,\textsc{Left}~15^{\circ},\,\textsc{Right}~15^{\circ},\,\textsc{Stop}\}$. Streaming end-to-end VLN is commonly formulated as a multi-turn dialogue with a Video-LLM: each user turn provides the current observation with the instruction and the retained history, and each assistant turn emits an action before the next observation arrives. A Video-LLM navigation policy comprises a vision encoder $\mathcal{V}$ that tokenizes the retained RGB observations, a projector $\phi$ that maps visual tokens into the LLM embedding space, and an autoregressive language model $\mathcal{F}_{\mathrm{LLM}}$ that processes the joint vision-language-action sequence. Let $\mathcal{W}_t$ denote the visual window retained at turn $t$, let $I_{\mathcal{W}_t}$ denote the corresponding RGB observations, which are encoded into $N_v(t)$ visual tokens, and let $P_t$ denote the $N_p(t)$ text tokens containing the instruction, dialogue markers, and previous actions. The hidden states for action generation are
\begin{equation}
H_t^{\mathrm{LLM}}
= \mathcal{F}_{\mathrm{LLM}}
\!\left(
\left[
\phi\!\left(\mathcal{V}(I_{\mathcal{W}_t})\right);
P_t
\right]
\right)
\in \R^{(N_v(t)+N_p(t))\times \dmodel}.
\label{eq:stream_hidden}
\end{equation}

\begin{figure}[!htbp]
    \centering
    \includegraphics[page=2,width=\textwidth,trim=380 595 366 580,clip]{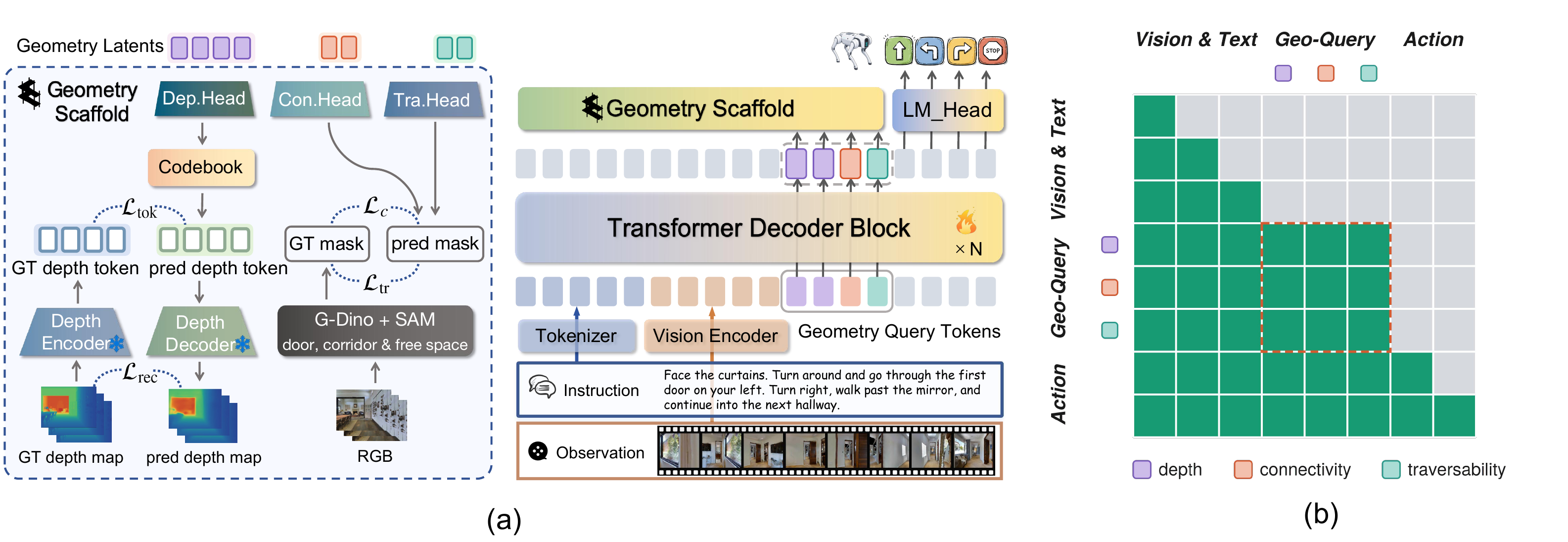}
    \vspace{-12pt}
    \caption{An overview of GeoScaffold. (a) Three geometric targets, namely depth tokens, a connectivity mask, and a traversability mask, supervise the hidden states of the query tokens through training-only reconstruction heads. (b) The hybrid attention mask, where the three query groups of one turn form a bidirectional block and all other entries stay causal.}
    \label{fig:pipeline}
    \vspace{-8pt}
\end{figure}
\subsection{Model Architecture}
\label{sec:model_arch}

Prior geometry-enhanced navigators consume geometry through dedicated spatial encoders and fusion modules that make it usable to the LLM \citep{zeng2025janusvln,yang2026gavln}. This machinery then stays resident at inference. We therefore reverse the interface. Rather than consuming geometry as an input, the navigator learns to predict it, compressing scene geometry into a small set of geometric latents that action decoding can attend to. The latents are supervised with discrete depth tokens and binary connectivity and traversability masks. Such classification-style targets are the kind of supervision the backbone is already trained on, in contrast to dense regression signals.

\paragraph{Geometry query tokens.}
We add $\numq{=}K_d{+}K_c{+}K_{\mathrm{tr}}$ new vocabulary tokens $\geotok{1},\dots,\geotok{\numq}$, partitioned into three groups dedicated to depth, connectivity, and traversability, respectively. These tokens are inserted at the end of every user turn, immediately before the assistant's action block, extending the input sequence of \cref{eq:stream_hidden} to
\begin{equation}
X_t=\bigl[\,\phi\!\left(\mathcal{V}(I_{\mathcal{W}_t})\right);\;P_t;\;\geotok{1};\cdots;\geotok{\numq}\,\bigr],
\qquad
H_t^{\mathrm{LLM}}=\mathcal{F}_{\mathrm{LLM}}\!\left(X_t\right)\in\R^{(N_v(t)+N_p(t)+\numq)\times\dmodel}.
\label{eq:geo_sequence}
\end{equation}
The query tokens are excluded from the language-modeling loss and never appear in generated text, while their input and output embeddings remain trainable. Because they precede the action block, subsequent action tokens can attend to them through the causal part of the attention mask. The hidden states at the query positions realize the geometric latents introduced above, carrying a compact geometric summary of the current observation into action decoding.

\paragraph{Hybrid attention for query tokens.}
Standard causal masking would impose an arbitrary left-to-right order on the $\numq$ query slots, which jointly summarize a single observation and carry no intrinsic ordering. We therefore unmask each turn's query group internally. Let $\mathbf{A}\in\{0,1\}^{L_t\times L_t}$ denote the attention mask over $X_t$ with $L_t=N_v(t)+N_p(t)+\numq$, and let $\mathcal{Q}^{(s)}$ denote the positions of the query tokens inserted at turn $s$. All non-query positions follow the standard causal rule, and for a query position $i\in\mathcal{Q}^{(s)}$ we additionally unmask its own group:
\begin{equation}
\mathbf{A}_{ij}=1
\quad\Longleftrightarrow\quad
j\le i \;\;\text{or}\;\; j\in\mathcal{Q}^{(s)}.
\label{eq:hybrid_mask}
\end{equation}
Query tokens of a turn attend to one another in both directions, and other positions stay causal.

\paragraph{Geometric supervision as a scaffold.}
At each supervised turn $t$, we build one target per type $g\in\mathcal{G}=\{d,c,\mathrm{tr}\}$, namely depth codes, a connectivity mask, and a traversability mask:
\begin{equation}
\tau_t^{d}=\vqenc(\depthmap_t)\in\{1,\dots,\numcode\}^{\numtok},\qquad
m_t^{c}=\operatorname{vec}(\mathcal{C}_t)\in\{0,1\}^{M},\qquad
m_t^{\mathrm{tr}}=\operatorname{vec}(\mathcal{T}_t)\in\{0,1\}^{M},
\label{eq:geo_targets}
\end{equation}
where $M=H_gW_g$. The depth target $\tau_t^d$ is the output of the frozen tokenizer, computed without gradients. The traversability map $\mathcal{T}_t$ marks free-space and floor-like regions through which the agent can move, while the connectivity map $\mathcal{C}_t$ marks passages relevant to route continuation, such as doors, corridors, and openings. Both maps are generated offline from the egocentric RGB frames with Grounding-DINO \citep{liu2024grounding} and SAM \citep{kirillov2023segment}.
\begin{figure}[!htbp]
    \centering
    \includegraphics[page=3,width=0.72\textwidth,trim=682 738 900 677,clip]{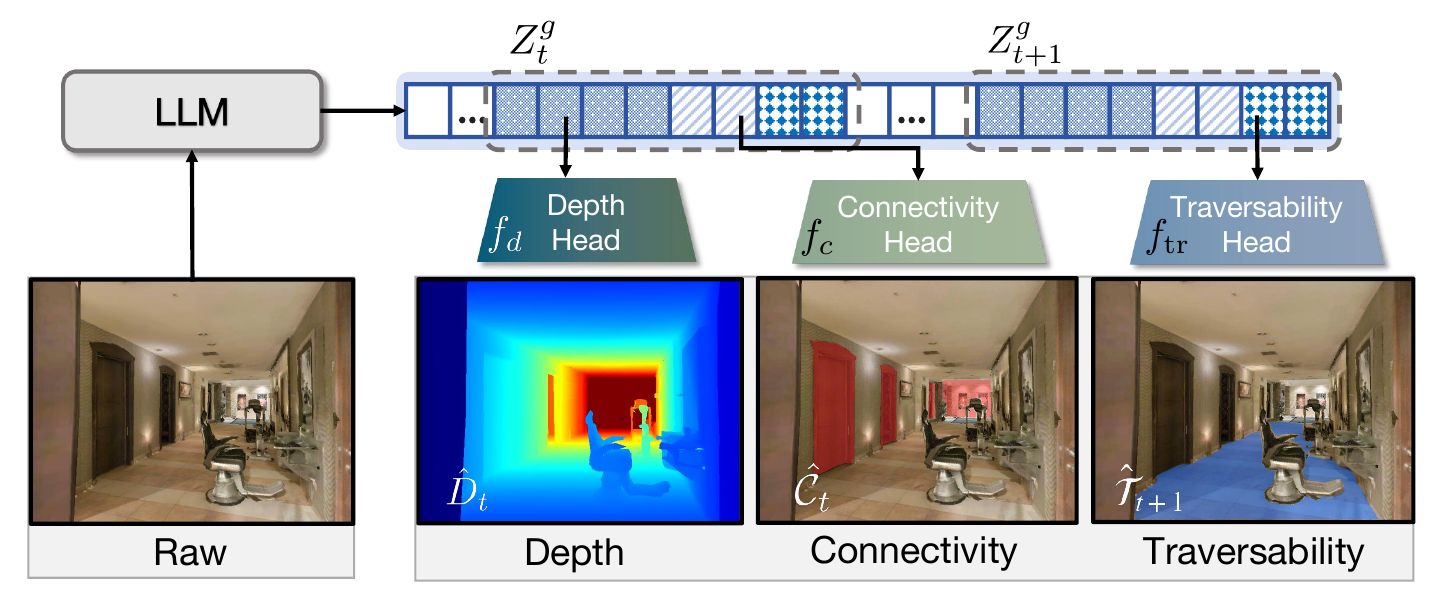}
    \caption{Depth, connectivity, and traversability targets decoded from the query-token interface. Traversability highlights free or floor-like regions, while connectivity emphasizes passages and openings relevant to route continuation.}
    \label{fig:geometric-targets}
\end{figure}

Let $\mathcal{Q}_g$ denote the set of query-token positions assigned to geometry type $g$, and let $K_g=|\mathcal{Q}_g|$. We gather the corresponding last-layer hidden states from $H_t^{\mathrm{LLM}}$:
\begin{equation}
Z_t^g = H_t^{\mathrm{LLM}}[\mathcal{Q}_g]\in\R^{K_g\times\dmodel},
\qquad g\in\mathcal{G}.
\label{eq:query_states}
\end{equation}
Each group of query states is flattened and decoded by a target-specific reconstruction head:
\begin{equation}
\hat{\tau}_t^{d}=f_d(\operatorname{vec}(Z_t^d))\in\R^{\numtok\times\numcode},\quad
\hat{m}_t^{c}=f_c(\operatorname{vec}(Z_t^c))\in\R^{M},\quad
\hat{m}_t^{\mathrm{tr}}=f_{\mathrm{tr}}(\operatorname{vec}(Z_t^{\mathrm{tr}}))\in\R^{M}.
    \label{eq:head}
\end{equation}
These are logits: $\hat{\tau}_t^{d}$ holds one logit per codebook entry at each depth-token position, while $\hat{m}_t^{c}$ and $\hat{m}_t^{\mathrm{tr}}$ hold one logit per pixel of the connectivity and traversability masks. We write $\hat{\mathcal{C}}_t$ and $\hat{\mathcal{T}}_t$ for these predictions reshaped back to $H_g\times W_g$. Thus, four query tokens jointly predict the $\numtok$ discrete VQ depth codes in a single forward pass, two query tokens predict the binary connectivity mask, and two query tokens predict the binary traversability mask. The depth codes are not added to the language vocabulary and are never generated autoregressively.

\paragraph{Removing the scaffold at inference.}
After training, the reconstruction heads and the depth tokenizer are removed. The deployed policy is the original backbone with the learned query-token embeddings in the prompt and the hybrid mask of \cref{eq:hybrid_mask}, a masking pattern that adds no parameters or computation. At inference the navigator receives only RGB, the instruction, and the dialogue history, and the acquired geometry reaches action decoding through the geometric latents at the query positions.

\subsection{Two-Stage Training Strategy}
\label{sec:training_strategy}

\paragraph{Stage I: depth tokenizer training.}
The first stage learns a discrete, decodable representation of per-frame geometry. We train a small depth VQ-VAE \citep{oord2017neural} offline on normalized depth maps from the navigation \emph{training} trajectories. The tokenizer is never exposed to evaluation environments. The convolutional encoder $\vqenc$ maps a depth map $\depthmap_t$ to a grid of $\numtok$ latents, each of which is quantized against a $\numcode$-entry codebook $\codebook$, yielding the depth code indices $\tau_t^d$. The convolutional decoder $\vqdec$ reconstructs the depth map from the corresponding code embeddings. The tokenizer is trained with the standard VQ-VAE objective, comprising reconstruction, codebook, and commitment terms. After this stage, $\vqenc$, $\vqdec$, and $\codebook$ are frozen. During policy training, $\vqenc$ serves only as a target generator, and $\vqdec$ serves only as a fixed differentiable renderer for the soft reconstruction loss.

\paragraph{Stage II: geometry-scaffolded policy training.}
The second stage fine-tunes the navigation policy with the query tokens and hybrid mask of \cref{sec:model_arch}. Alongside the unchanged imitation objective $\lact$, the reconstruction heads decode the query hidden states and are supervised against the frozen targets of \cref{eq:geo_targets} through three losses. The depth-token loss $\ltok$ is the mean cross-entropy over the $\numtok$ depth-token positions. To supervise decodable geometry beyond token identity, the reconstruction loss $\lrec$ renders the predicted code distribution through the frozen decoder and compares the result with the ground-truth depth map. The two losses are
\begin{equation}
\ltok = -\tfrac{1}{\numtok}\sum_{n=1}^{\numtok}\log\softmax\!\left(\hat{\tau}_t^{d,n}\right)_{\tau_t^{d,n}},\qquad
\lrec = \bigl\lVert \vqdec\bigl(\softmax(\hat{\tau}_t^{d})\,\codebook\bigr)-\depthmap_t \bigr\rVert_2^2,
\label{eq:depth_losses}
\end{equation}
where $\hat{\tau}_t^{d,n}\in\R^{\numcode}$ denotes the predicted code logits and $\tau_t^{d,n}$ the code index produced by the frozen tokenizer at depth-token position $n$. Gradients propagate through the frozen decoder to the reconstruction head and the Video-LLM, while the decoder parameters remain fixed. For each mask target $g\in\{c,\mathrm{tr}\}$, with target $m_t^g$ and predicted logits $\hat{m}_t^g$, we employ a binary mask reconstruction loss that combines pointwise classification with a soft region-overlap term:
\begin{equation}
\mathcal{L}_{g}
= \mathrm{BCE}\bigl(\hat{m}_t^g,\,m_t^g\bigr)
+ \lambda_{\mathrm{dice}}\,
\mathcal{L}_{\mathrm{Dice}}\bigl(\sigma(\hat{m}_t^g),\,m_t^g\bigr),
\label{eq:mask_loss}
\end{equation}
where $\sigma(\cdot)$ is the sigmoid. The objective adds the geometric reconstruction losses to the imitation loss:
\begin{equation}
\mathcal{L}
= \lact
+ \lambda_{\mathrm{tok}}\,\ltok
+ \lambda_{\mathrm{rec}}\,\lrec
+ \lambda_{c}\,\mathcal{L}_{c}
+ \lambda_{\mathrm{tr}}\,\mathcal{L}_{\mathrm{tr}}.
\label{eq:total}
\end{equation}

\section{Experiments}
\label{sec:experiments}

\subsection{Experimental Setup}
\label{sec:setup}

\paragraph{Datasets \& evaluation metrics.}
We train and evaluate in the VLN-CE setting \citep{krantz2020beyond}, where trajectories are executed as low-level actions in Matterport3D scenes \citep{chang2017matterport3d}. Our navigation-action data consists of oracle trajectories from R2R-CE and RxR-CE, extended with the extra navigation data and DAgger trajectories of the StreamVLN recipe \citep{wei2025streamvln}. Every trajectory frame is additionally paired with generated depth, connectivity, and traversability annotations (\cref{sec:model_arch}), which serve only as training-time reconstruction targets. We report the standard VLN-CE metrics, namely Navigation Error (NE), Success Rate (SR), Oracle Success Rate (OS), Success weighted by Path Length (SPL), and normalized Dynamic Time Warping (nDTW), among which SR and SPL are the widely adopted primary measures of task completion and efficiency.

\paragraph{Implementation details.}
We adopt the streaming VLN formulation of StreamVLN \citep{wei2025streamvln}, a multi-turn dialogue with slow-fast visual context, and instantiate it on Qwen3-VL-4B \citep{qwen3vl}, decoding one action per turn. The geometry branch uses $\numq{=}8$ query tokens ($K_d{=}4$, $K_c{=}K_{\mathrm{tr}}{=}2$) read from the last hidden layer, and the depth tokenizer is a 0.74M-parameter VQ-VAE with $\numtok{=}100$ tokens over a $\numcode{=}128$-entry codebook. Throughout, a policy without GeoScaffold denotes the same backbone trained with the identical recipe and data but without the query tokens, reconstruction heads, and geometric losses. Training takes approximately 1500 A100 GPU hours for the policy and around 50 for the VQ-VAE.

\subsection{Main Results}
\label{sec:results}
\paragraph{Results on VLN-CE benchmarks.}
\Cref{tab:sota} compares GeoScaffold with published systems on R2R-CE and RxR-CE Val-Unseen. Rows marked with $^*$ use only R2R-CE and RxR-CE for navigation training, whereas the final block draws on large-scale additional navigation data such as DAgger trajectories and augmented corpora.
Two observations stand out. First, geometric supervision is data-efficient. Trained on VLN-CE data alone, GeoScaffold-4B$^*$ reaches 55.3\% SR and 51.0\% SPL on R2R-CE, ahead of JanusVLN-7B$^*$ under the same data, and in SR and SPL it already exceeds NaVILA-7B on both benchmarks and matches or exceeds StreamVLN-7B and InternVLA-N1 on RxR-CE, although these use more navigation data. With the additional data, GeoScaffold-4B reaches 64.8\% SR and 60.5\% SPL on R2R-CE, exceeding JanusVLN-7B by 4.3 and 3.7 points in SR and SPL and the RGB-D policy GA-VLN-7B by 3.8 and 5.3 points while observing only RGB. Second, GeoScaffold-4B achieves higher SR and SPL on both benchmarks than the 7B VLA systems listed in \cref{tab:sota}. On RxR-CE, whose instructions are longer and more fine-grained, it reaches 59.6\% SR, 52.2\% SPL, and 66.3 nDTW and improves over NavFoM-7B by 2.2 and 2.8 points in SR and SPL.

\begin{table}[!htbp]
\centering
\caption{Comparison with published systems on R2R-CE and RxR-CE Val-Unseen.
Variants marked with $^*$ use only R2R-CE and RxR-CE for navigation
training. StreamVLN$^\dagger$ uses EnvDrop-augmented navigation data,
while NaVILA$^\dagger$ excludes human touring videos.}
\label{tab:sota}
\setlength{\tabcolsep}{1.8pt}
\renewcommand{\arraystretch}{1.15}
\footnotesize
\begin{tabular}{lcccccccccccc}
\toprule
\multirow{2}{*}{Method} & \multicolumn{4}{c}{Observation} & \multicolumn{4}{c}{R2R-CE Val-Unseen} & \multicolumn{4}{c}{RxR-CE Val-Unseen} \\
\cmidrule(lr){2-5}\cmidrule(lr){6-9}\cmidrule(lr){10-13}
 & {\scriptsize Pano.} & {\scriptsize Odo.} & {\scriptsize Depth} & {\scriptsize S.RGB} & {\scriptsize NE$\downarrow$} & {\scriptsize OS$\uparrow$} & {\scriptsize SR$\uparrow$} & {\scriptsize SPL$\uparrow$} & {\scriptsize NE$\downarrow$} & {\scriptsize SR$\uparrow$} & {\scriptsize SPL$\uparrow$} & {\scriptsize nDTW$\uparrow$} \\
\midrule
CMA \citep{hong2022bridging} & \checkmark & \checkmark & \checkmark & & 6.20 & 52.0 & 41.0 & 36.0 & 8.76 & 26.5 & 22.1 & 47.0 \\
Sim2Sim \citep{krantz2022sim2sim} & \checkmark & \checkmark & \checkmark & & 6.07 & 52.0 & 43.0 & 36.0 & -- & -- & -- & -- \\
VLN$\circlearrowright$BERT \citep{hong2022bridging} & \checkmark & \checkmark & \checkmark & & 5.74 & 53.0 & 44.0 & 39.0 & 8.98 & 27.0 & 22.6 & 46.7 \\
GridMM \citep{wang2023gridmm} & \checkmark & \checkmark & \checkmark & & 5.11 & 61.0 & 49.0 & 41.0 & -- & -- & -- & -- \\
Reborn \citep{an2022reborn} & \checkmark & \checkmark & \checkmark & & 5.40 & 57.0 & 50.0 & 46.0 & 5.98 & 48.6 & 42.0 & 63.3 \\
\midrule
LAW \citep{raychaudhuri2021law} & & \checkmark & \checkmark & \checkmark & 6.83 & 44.0 & 35.0 & 31.0 & 10.90 & 8.0 & 8.0 & 38.0 \\
Seq2Seq \citep{krantz2020beyond} & & & \checkmark & \checkmark & 7.77 & 37.0 & 25.0 & 22.0 & 12.10 & 13.9 & 11.9 & 30.8 \\
NaVid-4D \citep{liu2025navid4d} & & & \checkmark & \checkmark & 5.99 & 55.7 & 43.8 & 37.1 & -- & -- & -- & -- \\
NavMorph \citep{yao2025navmorph} & & & \checkmark & \checkmark & 5.75 & 56.9 & 47.9 & 33.2 & -- & -- & -- & -- \\
GA-VLN-7B \citep{yang2026gavln} & & \checkmark & \checkmark & \checkmark & 4.80 & 67.6 & 61.0 & 55.2 & 5.88 & 55.4 & 45.2 & -- \\
\midrule
NaVid-7B \citep{zhang2024navid} & & & & \checkmark & 5.47 & 49.1 & 37.4 & 35.9 & -- & -- & -- & -- \\
StreamVLN-7B$^\dagger$ \citep{wei2025streamvln} & & & & \checkmark & 6.05 & 53.8 & 45.5 & 41.6 & -- & -- & -- & -- \\
Uni-NaVid-7B \citep{zhang2025uninavid} & & & & \checkmark & 5.58 & 53.3 & 47.0 & 42.7 & 6.24 & 48.7 & 40.9 & -- \\
NaVILA-7B$^\dagger$ \citep{cheng2024navila} & & & & \checkmark & 5.37 & 57.6 & 49.7 & 45.5 & -- & -- & -- & -- \\
JanusVLN-7B$^*$ \citep{zeng2025janusvln} & & & & \checkmark & 5.17 & 58.0 & 52.8 & 49.2 & 6.46 & 51.4 & 44.3 & 59.1 \\
\rowcolor{gray!25} \textbf{GeoScaffold-4B$^*$ (Ours)} & & & & \checkmark & \textbf{5.07} & \textbf{61.9} & \textbf{55.3} & \textbf{51.0} & \textbf{5.82} & \textbf{53.2} & \textbf{46.2} & \textbf{59.8} \\
\midrule
NaVILA-7B \citep{cheng2024navila} & & & & \checkmark & 5.22 & 62.5 & 54.0 & 49.0 & 6.77 & 49.3 & 44.0 & 58.8 \\
StreamVLN-7B \citep{wei2025streamvln} & & & & \checkmark & 4.98 & 64.2 & 56.9 & 51.9 & 6.22 & 52.9 & 46.0 & 61.9 \\
InternVLA-N1 \citep{internrobotics2025internvla} & & & & \checkmark & 4.89 & 60.6 & 55.4 & 52.1 & 6.41 & 49.5 & 41.8 & 62.6 \\
NavFoM-7B \citep{zhang2025navfom} & & & & \checkmark & 5.01 & 64.9 & 56.2 & 51.2 & 5.51 & 57.4 & 49.4 & 60.2 \\
JanusVLN-7B \citep{zeng2025janusvln} & & & & \checkmark & 4.78 & 65.2 & 60.5 & 56.8 & 6.06 & 56.2 & 47.5 & 62.1 \\
\rowcolor{gray!25} \textbf{GeoScaffold-4B (Ours)} & & & & \checkmark & \textbf{4.17} & \textbf{72.1} & \textbf{64.8} & \textbf{60.5} & \textbf{4.51} & \textbf{59.6} & \textbf{52.2} & \textbf{66.3} \\
\bottomrule
\end{tabular}
\end{table}

\paragraph{Inference efficiency against geometry-enhanced methods.}
\begin{wrapfigure}[6]{r}{0.48\textwidth}
\vspace{-16pt}
\centering
\resizebox{\linewidth}{!}{\begin{tikzpicture}[font=\sffamily\scriptsize,yscale=0.8]
  \definecolor{ours}{HTML}{D55E00}
  \definecolor{otherc}{HTML}{5F6B7A}
  \definecolor{otherd}{HTML}{8A96A6}
  \definecolor{vggtc}{HTML}{353B45}

  \draw[gray!45,line width=0.5pt] (0,0.30) -- (0,1.86);

  \filldraw[ours,rounded corners=1pt] (0,1.47) rectangle (0.38,1.77);
  \node[ours,anchor=west,font=\sffamily\scriptsize\bfseries] at (0.48,1.62) {81\,ms};
  \node[ours,anchor=east,font=\sffamily\scriptsize\bfseries] at (-0.12,1.62) {GeoScaffold-4B};

  \filldraw[otherd,rounded corners=1pt] (0,0.93) rectangle (1.80,1.23);
  \node[otherd!80!black,anchor=west,font=\sffamily\scriptsize\bfseries] at (1.90,1.08) {750\,ms};
  \node[otherd!80!black,anchor=east,font=\sffamily\scriptsize\bfseries] at (-0.12,1.08) {GA-VLN};

  \filldraw[otherc,rounded corners=1pt] (0,0.39) rectangle (1.95,0.69);
  \filldraw[vggtc,rounded corners=1pt] (2.16,0.39) rectangle (2.80,0.69);
  \draw[white,line width=3pt] (1.99,0.33) -- (2.13,0.75);
  \draw[gray!60,line width=0.6pt] (1.96,0.33) -- (2.10,0.75);
  \draw[gray!60,line width=0.6pt] (2.02,0.33) -- (2.16,0.75);
  \node[white,anchor=center,font=\sffamily\tiny] at (2.48,0.54) {VGGT};
  \node[vggtc,anchor=north,font=\sffamily\tiny\bfseries] at (2.48,0.32) {400\,ms};
  \node[otherc,anchor=west,font=\sffamily\scriptsize\bfseries] at (2.90,0.54) {1160\,ms};
  \node[otherc,anchor=east,font=\sffamily\scriptsize\bfseries] at (-0.12,0.54) {JanusVLN};
\end{tikzpicture}}
\vspace{-22pt}
\caption{Average per-frame latency.}
\label{fig:latency}
\vspace{-10pt}
\end{wrapfigure}
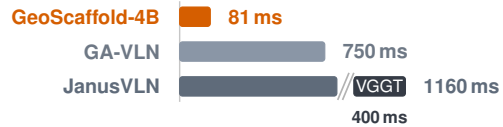

We benchmark inference on a single NVIDIA RTX 4090 and report the average wall-clock latency to process one incoming frame and emit its action. All measurements are taken in the steady state, after the streaming context has filled the sliding window and the per-frame cost has stabilized. As shown in \cref{fig:latency}, GeoScaffold-4B processes a frame in 81\,ms. GA-VLN \citep{yang2026gavln}, which lifts RGB-D observations into a geometry-aware BEV map enriched with 3D foundation-model features, requires 750\,ms, and JanusVLN \citep{zeng2025janusvln}, which runs a resident VGGT 3D encoder at every step, requires 1{,}160\,ms. The two geometry-enhanced policies are therefore $9\times$ to $14\times$ slower than GeoScaffold-4B. The VGGT encoder alone accounts for 400\,ms, almost $5\times$ the per-frame inference of GeoScaffold-4B. Training-time geometry thus buys spatial competence without the standing inference cost that resident geometry modules impose, and GeoScaffold-4B sustains control rates above 12\,Hz on a single consumer GPU.

\paragraph{Backbone generality.}
Because GeoScaffold changes only the supervision and modifies no architecture, sensor, or action interface, porting it to a new backbone amounts to re-running training. We apply the identical recipe to three backbones from three model families, namely LLaVA-Video-7B, Qwen2.5-VL-3B, and Qwen3-VL-4B, and train each of them without and with GeoScaffold on R2R-CE and RxR-CE only, the data configuration used for all ablations. \Cref{tab:backbone} reports the paired results. GeoScaffold improves every backbone on all four metrics. SR rises by 5.5 to 7.9 points, SPL by 6.4 to 8.7 points, and NE falls by 0.5 to 0.8\,m. The gains hold across three model families and model sizes from 3B to 7B without any change to the recipe, so GeoScaffold transfers to a new backbone as a drop-in training procedure.

\begin{table}[!htbp]
\centering
\caption{Backbone generality. Each backbone is trained with the identical recipe without and with GeoScaffold on R2R-CE and RxR-CE only, and evaluated on R2R-CE Val-Unseen.}
\label{tab:backbone}
\small
\setlength{\tabcolsep}{6pt}
\begin{tabular}{l|cccc|cccc|c}
\toprule
\multirow{2}{*}{Backbone} & \multicolumn{4}{c|}{w/o GeoScaffold} & \multicolumn{4}{c|}{w/ GeoScaffold} & \multirow{2}{*}{$\Delta$SR} \\
& NE$\downarrow$ & OS$\uparrow$ & SR$\uparrow$ & SPL$\uparrow$ & NE$\downarrow$ & OS$\uparrow$ & SR$\uparrow$ & SPL$\uparrow$ & \\
\midrule
LLaVA-Video-7B & 6.15 & 50.2 & 39.4 & 37.1 & 5.61 & 56.8 & 47.3 & 45.8 & $+$7.9 \\
Qwen2.5-VL-3B & 5.96 & 53.7 & 46.4 & 41.6 & 5.14 & 58.5 & 51.9 & 49.2 & $+$5.5 \\
Qwen3-VL-4B & 5.77 & 56.2 & 48.8 & 44.6 & 5.07 & 61.9 & 55.3 & 51.0 & $+$6.5 \\
\bottomrule
\end{tabular}
\end{table}

\subsection{Qualitative Results}
\label{sec:realworld}

\begin{figure}[!htbp]
    \centering
    \vspace{-4pt}
    \includegraphics[page=4,width=0.98\textwidth,trim=447 431 463 809,clip]{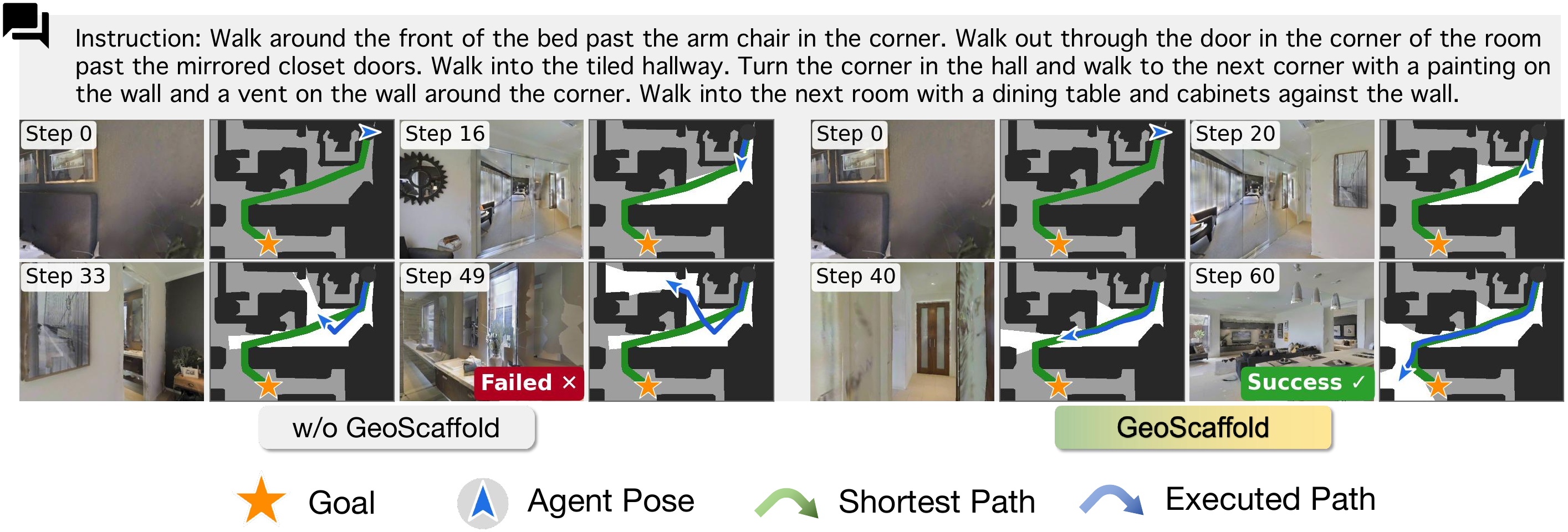}
    \vspace{-4pt}
    \caption{Simulated qualitative comparison on VLN-CE.}
    \label{fig:sim_qual}
    \vspace{-10pt}
\end{figure}

\paragraph{Simulation.}
\Cref{fig:sim_qual} compares closed-loop rollouts with and without GeoScaffold on an R2R-CE Val-Unseen episode from the same start pose and instruction. Without GeoScaffold the policy leaves the hallway too early and stops prematurely in an adjacent room. With GeoScaffold the policy follows the hallway around the corner and stops only in the described dining room. The rollouts thus show a policy that is more sensitive to spatial structure, to which openings continue the route and to how far it still has to go.

\begin{figure}[!htbp]
    \centering
    \vspace{-4pt}
    \includegraphics[page=5,width=0.98\textwidth,trim=313 750 291 385,clip]{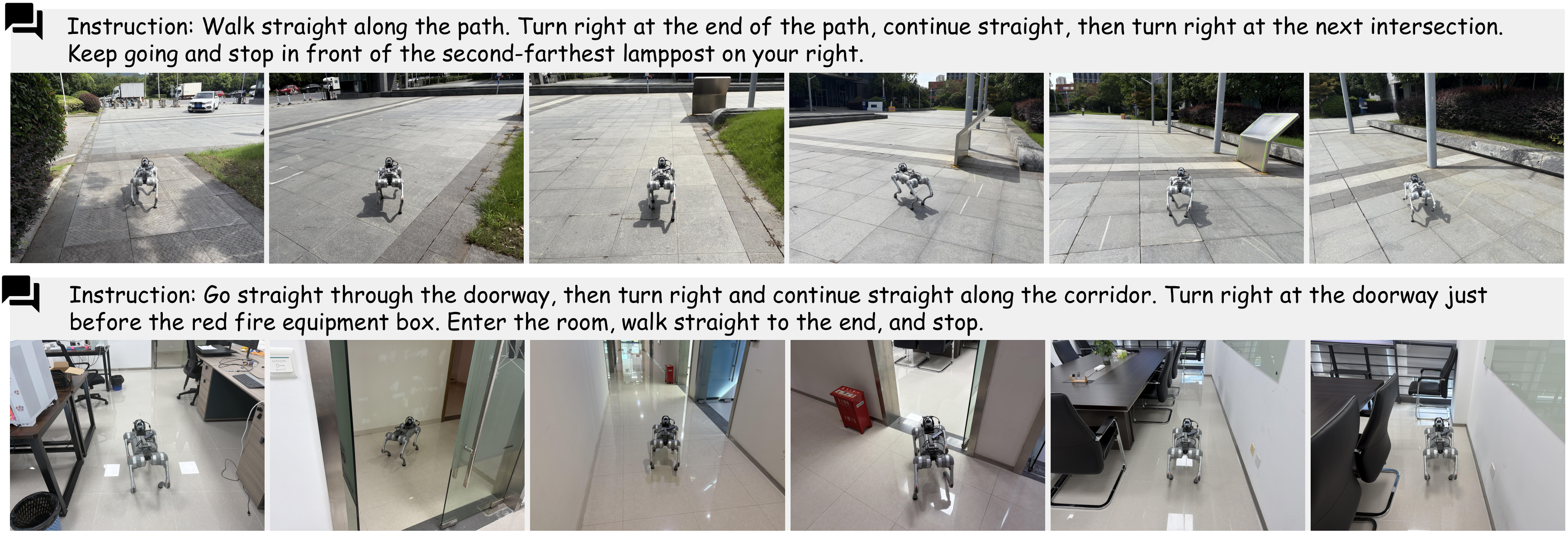}
    \vspace{-8pt}
    \caption{Zero-shot real-world qualitative results.}
    \label{fig:realworld}
    \vspace{-8pt}
\end{figure}

\paragraph{Real-world deployment.}
Policies with a resident depth or 3D encoder are costly to run on a robot and are often moved to off-board servers, where remote communication and asynchronous inference add latency and lower the control frequency. GeoScaffold carries no such cost, and the Qwen3-VL-4B policy can run locally on an NVIDIA Jetson edge device from RGB and language alone. We deploy it on a Unitree Go2 Edu quadruped with an Intel RealSense D435i camera, using only its RGB stream, and \cref{fig:realworld} shows two zero-shot episodes. The geometry-demanding instructions of our real-world experiments stress three forms of spatial understanding: depth-aware reasoning toward distant landmarks such as a street lamp, traversability reasoning when objects partially block the route, and connectivity reasoning in doorway and porch-like scenes. GeoScaffold completes both episodes and raises the success rate on geometry-demanding instructions over a larger instruction set, indicating that the spatial reasoning instilled by training-time supervision transfers to the real world.

\subsection{Ablation Study}
\label{sec:ablation}

Unless otherwise stated, all ablations in this section are trained without external data and evaluated on the R2R-CE Val-Unseen benchmark.

\paragraph{Supervision-target composition.}
\Cref{tab:abl_targets} ablates the three supervision targets reconstructed by the full model (\cref{sec:model_arch}) by removing one at a time. Depth is the dominant signal. Removing depth-token supervision lowers SR from 55.3 to 51.1, the largest single drop, with consistent drops in NE, OS, and SPL. The two masks contribute smaller but clear gains on top of depth: removing traversability costs 2.2 SR and removing connectivity 2.6 SR. Every variant degrades on all four metrics, so the targets are complementary rather than redundant. The masks add passage and free-space cues that depth alone does not capture. We therefore adopt all three targets throughout.

\begin{table}[H]
\vspace{-8pt}
\begin{minipage}[t]{0.44\textwidth}
\centering
\caption{Ablation on supervision-target composition.}
\label{tab:abl_targets}
\footnotesize
\setlength{\tabcolsep}{3pt}
\begin{tabular}{lcccc}
\toprule
Variant & NE$\downarrow$ & OS$\uparrow$ & SR$\uparrow$ & SPL$\uparrow$ \\
\midrule
\textbf{GeoScaffold} & 5.07 & 61.9 & 55.3 & 51.0 \\
\quad w/o depth & 5.25 & 57.4 & 51.1 & 47.1 \\
\quad w/o traversability & 5.23 & 58.7 & 53.1 & 49.2 \\
\quad w/o connectivity & 5.18 & 58.5 & 52.7 & 48.2 \\
\bottomrule
\end{tabular}
\end{minipage}\hfill
\begin{minipage}[t]{0.54\textwidth}
\centering
\caption{Ablation on depth representation and losses.}
\label{tab:abl_objective}
\footnotesize
\setlength{\tabcolsep}{3pt}
\begin{tabular}{lcccc}
\toprule
Depth target / loss & NE$\downarrow$ & OS$\uparrow$ & SR$\uparrow$ & SPL$\uparrow$ \\
\midrule
\textbf{VQ-VAE, $\ltok{+}\lrec$ (Ours)} & 5.11 & 59.4 & 52.3 & 48.8 \\
\quad w/o $\ltok$ ($\lrec$ only) & 5.93 & 55.2 & 48.5 & 43.7 \\
\quad w/o $\lrec$ ($\ltok$ only) & 5.19 & 57.8 & 51.0 & 46.4 \\
Continuous VAE latent & 5.35 & 56.6 & 50.9 & 45.9 \\
Pixel-space regression & 5.66 & 55.8 & 49.3 & 44.6 \\
\bottomrule
\end{tabular}
\end{minipage}
\vspace{-8pt}
\end{table}

\paragraph{Depth representation and reconstruction loss.}
\Cref{tab:abl_objective} varies how the depth target is represented and supervised, with depth as the only supervision target. Discrete token classification carries most of the signal. Dropping $\ltok$ costs 3.8 SR, whereas dropping $\lrec$ costs 1.3, and the two losses together remain best. Discreteness also matters. Replacing the VQ codes with a capacity-matched continuous latent drops 1.4 SR, and pixel-space depth regression costs 3.0 SR. Compact, discrete, decodable targets are thus the right target for supervising a token-based policy.

\paragraph{Hidden-state source for reconstruction.}
\Cref{tab:abl_hidden_source} varies which positions the reconstruction heads read from, with targets and losses unchanged. Supervising image hidden states leads to the largest SR drop (12.1 points), possibly because forcing visual features to support geometric reconstruction degrades their quality for navigation. Supervising action hidden states costs 3.5 SR, as the reconstruction gradient competes with the imitation objective at the positions that generate actions. Query tokens without reconstruction fall to 47.7 SR, so the gains come from the geometric objective rather than from the extra tokens. In short, query tokens work best because they absorb the geometric supervision at positions that the backbone does not otherwise use for perception or action.

\begin{table}[H]
\vspace{-8pt}
\begin{minipage}[t]{0.46\textwidth}
\centering
\caption{Ablation on the hidden-state source.\vphantom{q}}
\label{tab:abl_hidden_source}
\footnotesize
\setlength{\tabcolsep}{3pt}
\begin{tabular}{lcccc}
\toprule
Hidden-state source & NE$\downarrow$ & OS$\uparrow$ & SR$\uparrow$ & SPL$\uparrow$ \\
\midrule
Image hidden states & 6.78 & 50.6 & 43.2 & 38.5 \\
Action hidden states & 5.23 & 58.4 & 51.8 & 47.9 \\
\textbf{Query tokens (Ours)} & 5.07 & 61.9 & 55.3 & 51.0 \\
\quad w/o reconstruction & 6.06 & 54.8 & 47.7 & 43.3 \\
\bottomrule
\end{tabular}
\end{minipage}\hfill
\begin{minipage}[t]{0.52\textwidth}
\centering
\caption{Ablation on the query-token attention mask.}
\label{tab:abl_mask}
\footnotesize
\setlength{\tabcolsep}{3pt}
\begin{tabular}{lcccc}
\toprule
Attention mask & NE$\downarrow$ & OS$\uparrow$ & SR$\uparrow$ & SPL$\uparrow$ \\
\midrule
\textbf{Hybrid mask (Ours)} & 5.07 & 61.9 & 55.3 & 51.0 \\
\quad latents blocked at test & 5.17 & 58.9 & 52.8 & 47.7 \\
Causal mask & 5.15 & 59.6 & 53.9 & 48.1 \\
Causal, reversed order & 5.20 & 59.8 & 53.5 & 48.7 \\
\bottomrule
\end{tabular}
\end{minipage}
\vspace{-8pt}
\end{table}

\paragraph{Attention design and latent utilization.}
\Cref{tab:abl_mask} isolates the attention design. Replacing the hybrid mask with a causal mask over the query tokens costs 1.4 SR. Reversing the query order in the causal variant changes SR only slightly, from 53.9 to 53.5, suggesting limited sensitivity to query order. The hybrid mask achieves the best performance. Blocking attention to the latents at test time, without retraining, lowers SR from 55.3 to 52.8 and SPL from 51.0 to 47.7 with the weights untouched, so action decoding does read the geometric latents at inference. The blocked variant still outperforms the policy without GeoScaffold (48.8 SR, \cref{tab:backbone}), indicating that geometry is also internalized into the model's own representations.

\section{Conclusion}
\label{sec:conclusion}

We presented GeoScaffold, a training-time supervision framework that internalizes geometry into streaming VLN policy representations. By reconstructing depth, connectivity, and traversability from dedicated query states, it learns compact geometric latents for action decoding. The auxiliary modules are discarded after training, leaving a policy that navigates from RGB and language without a resident geometry encoder. Experiments demonstrate navigation gains on R2R-CE and RxR-CE, across multiple backbones, and in real-world deployment. The 4B policy averages 81\,ms per frame on an RTX 4090 and also runs on board a Jetson device. These findings support geometric reconstruction as a practical training signal for efficient, spatially aware navigation.

\end{document}